\documentclass{article}

\usepackage{microtype}
\usepackage{graphicx}
\usepackage{subfigure}
\usepackage{booktabs} % for professional tables

\usepackage{amsmath}
\usepackage{amsfonts}

\usepackage{hyperref}

\usepackage[accepted]{icml2020}

\icmltitlerunning{Reinforcement Learning Enhanced Quantum-inspired Algorithm for Combinatorial Optimization}

\begin{document}

\twocolumn[
\icmltitle{Reinforcement Learning Enhanced Quantum-inspired Algorithm \\ for Combinatorial Optimization}



\icmlsetsymbol{equal}{*}

\begin{icmlauthorlist}
\icmlauthor{Dmitrii Beloborodov}{rqc}
\icmlauthor{A. E. Ulanov}{rqc}
\icmlauthor{Jakob N. Foerster}{cs}
\icmlauthor{Shimon Whiteson}{cs}
\icmlauthor{A. I. Lvovsky}{rqc,ph}
\end{icmlauthorlist}

\icmlaffiliation{rqc}{Russian Quantum Center, Moscow, Russia}
\icmlaffiliation{cs}{Department of Computer Science, University of Oxford, Oxford, United Kingdom}
\icmlaffiliation{ph}{Department of Physics, University of Oxford, Oxford, United Kingdom}

\icmlcorrespondingauthor{Dmitrii Beloborodov}{\mbox{dmitribeloborodov@yandex.ru}}

\icmlkeywords{Combinatorial Optimization, Reinforcement Learning, Ising Model, Maximum Cut}

\vskip 0.3in
]



\printAffiliationsAndNotice{\icmlEqualContribution} % otherwise use the standard text.

\begin{abstract}
Quantum hardware and quantum-inspired algorithms are becoming increasingly popular for combinatorial optimization. However, these algorithms may require careful hyperparameter tuning for each problem instance. We use a reinforcement learning agent in conjunction with a quantum-inspired algorithm to solve the Ising energy minimization problem, which is equivalent to the Maximum Cut problem. The agent controls the algorithm by tuning one of its parameters with the goal of improving recently seen solutions. We propose a new Rescaled Ranked Reward (R3) method that enables a stable single-player version of self-play training and helps the agent escape local optima. The training on any problem instance can be accelerated by applying transfer learning from an agent trained on randomly generated problems. Our approach allows sampling high quality solutions to the Ising problem with high probability and outperforms both baseline heuristics and a black-box hyperparameter optimization approach.
\end{abstract}

\section{Introduction}
\label{intro}

Many important real-world combinatorial problems can be mapped to the Ising model, ranging from portfolio optimization \cite{portfolio1, portfolio2} to protein folding \cite{protein}. The Ising model describes the pairwise interaction of binary particles and assigns some cost function (energy) to each particle configuration. The Ising problem consists in finding binary strings that minimize the energy. It is a quadratic unconstrained optimization task over the discrete $\{\pm1\}^n$ domain and equivalent to the Max-Cut problem from graph theory.

There are multiple methods for solving the Ising and Max-Cut problems. Classic algorithms include heuristics performing local search in the solution space, like breakout local search \cite{bls} and simulated annealing \cite{annealing}. For many combinatorial problems, commercial solvers are available, including Gurobi  \cite{gurobi} and CPLEX \cite{cplex}.

An entirely different approach is to use a quantum physical system with its energy function similar to the optimization objective, and then anneal this system towards its ground state --- the lowest energy state. Devices utilizing this method include the coherent Ising machine (CIM) \cite{cim1, cim2} and the quantum superconducting annealer manufactured by D-Wave Systems \cite{dwave}. For example, in CIM, pulses of light circulate in a lossy optical fiber loop containing a parametric amplifier. In each round trip, a classical controller modulates these pulses according to the parameters of the Ising problem and the measured amplitudes of other pulses.

Quantum technology does not yet compete with classical computation systems in terms of both problem size and solution quality. However, it has inspired a family of new classical optimisation algorithms that perform well in comparison with existing ones \cite{nmfa, destab}. An example is a simulation of CIM, known as the SimCIM algorithm \cite{sim}. SimCIM reformulates the Ising model as a continuous constrained optimization problem and solves it with iterative gradient-based optimization, with each iteration corresponding to a roundtrip of the optical pulses through the fiber loop. SimCIM was implemented on computers equipped with consumer GPUs and outperformed CIM in both solution quality and computation time \cite{sim}. It has since been applied as a Boltzmann sampler to train general Boltzmann machines and for applications in statistical physics \cite{ulanov2019}. However, both SimCIM and CIM require parameter tuning for each problem instance to obtain the best results. One of the SimCIM parameters, the combined coefficients of linear gain and loss (which can be interpreted as a dynamic regularization coefficient), needs to be varied as a function of the iteration number. As a result, the use of classic hyperparameter optimization approaches \cite{hpo} is limited, since most methods assume a small number of continuous or discrete parameters. 

To automate parameter tuning in a flexible way, we use a reinforcement learning agent to control the regularization (gain-loss) function of SimCIM during the optimization process. An important feature of the Ising problem is the presence of multiple local optima whose energy is only slightly higher than the global minimum, but the associated bit configuration is significantly different.  To address this issue, we propose Rescaled Ranked Reward (R3), a modification of Ranked Reward (R2) \cite{r2}. In this approach, we assign reward to the agent depending on how its current score compares to scores obtained in recent trials, and thus enable self-play training for a single-player environment. Rescaled Ranked Reward  ensures the agent is motivated to keep discovering better solutions, without destabilizing the training process. 

We demonstrate that the convergence speed noticeably improves if we apply policy transfer from an agent pre-trained on randomly generated problems to the unseen target problem. This transfer learning is facilitated by feature-wise linear modulation (FiLM) \cite{film} with the features extracted from the general parameters of the problem at hand.

Our approach allows us to find the best solutions with higher probability than SimCIM with a regularization function that changes linearly or according to a hyperbolic tangent function with manually tuned parameters (which is our benchmark for the human level performance). It also outperforms CMA-ES \cite{cmaes}, one of the most powerful black-box algorithms for  hyperparameter optimization.\footnote{The code is available at \url{https://github.com/BeloborodovDS/SIMCIM-RL}.}

\section{Background}

% Preable: Warning - detailed understanding is not required to follow the main ideas 

%This section describes in detail the Ising combinatorial problem, as well as the SimCIM algorithm: it is provided for interested readers. 

\subsection{Combinatorial optimization}

The Ising problem is a discrete energy minimization problem:

\begin{equation}
\label{eq:qubo}
\begin{cases}
- \boldsymbol{x}^T J \boldsymbol{x} \rightarrow \min_{\boldsymbol{x}}, \\
 J \in \mathbb{R}^{n \times n}, \ J^T = J, \ \  \boldsymbol{x} \in \{\pm 1\}^n
\end{cases}
\end{equation}

Here $\boldsymbol{x}$ is a vector of $n$ binary variables that can have values $\pm 1,$ and $J$ is a symmetric problem matrix that describes pairwise interactions between them. This problem is NP-hard \cite{nphard}. It is equivalent to the Max-cut problem: 

\begin{equation}
\label{eq:maxcut}
\begin{cases}
\mathcal{C}(J,\boldsymbol{x}) = \frac{1}{4}(\boldsymbol{x}^T J \boldsymbol{x} - \sum_{ij}J_{ij}) \rightarrow \max_{\boldsymbol{x}}, \\
 J \in \mathbb{R}^{n \times n}, \ J^T = J, \ \  \boldsymbol{x} \in \{\pm 1\}^n
\end{cases}
\end{equation}

This problem can be interpreted as a task of dividing a set of $n$ nodes of a weighted graph into two subsets, such that the sum of edge weights connecting these subsets $\mathcal{C}(J, \boldsymbol{x})$ is maximized. In this interpretation, the problem matrix $J$ is the adjacency matrix of the graph, and binary variables $\boldsymbol{x}$ denote the choice of the subset for each node. The optimization objective $\mathcal{C}(J, \boldsymbol{x})$ is called the \emph{cut value} (higher is better); in this paper we use it to evaluate our algorithm and compare it to benchmarks. 

\subsection{SimCIM algorithm}

SimCIM \cite{sim} solves the discrete problem (\ref{eq:qubo}) by replacing $\boldsymbol{x}$ with a vector $\boldsymbol{c}$ of $n$ continuous variables bounded in $[-1,1]$:

\begin{equation}
	\label{eq:rel}
	\begin{cases}
	- \boldsymbol{c}^T J \boldsymbol{c} + p \boldsymbol{c}^T \boldsymbol{c} \rightarrow \min_{\boldsymbol{c}}, \\
	J \in \mathbb{R}^{n \times n}, \ J^T = J, \ \boldsymbol{c} \in [-1,1]^n
	\end{cases}
\end{equation}
In addition to the Ising term, the SimCIM loss function contains a regularization term parametrized by the real scalar  $p$, which corresponds to the combined gain and linear loss coefficients in the CIM \cite{sim}; its value is allowed to change during the optimization process. This change does not have to be smooth or even continuous, however $p$ should  generally decrease with time to ensure that the loss function \eqref{eq:rel} approximates the continuous Ising loss function by the end of the iterations. 

The problem (\ref{eq:rel}) is solved via gradient descent. We initialize the vector $\boldsymbol c$ of problem variables according to $\boldsymbol{c}_1 = \{0\}^n$. At each  iteration $t$, we compute the noisy  anti-gradient of the loss function 
$$\boldsymbol{g}_{t} = \mu (J \boldsymbol{c}_t - p_t \boldsymbol{c}_t) + \sigma \boldsymbol{\varepsilon}_t,$$
where $\mu$ is the learning rate,  $p_t$ is the regularization coefficient,  $\boldsymbol{\varepsilon}_{t}$ is a vector of random samples from the standard normal distribution and $\sigma$ is the noise amplitude. Then we calculate the update vector  $\boldsymbol{m}_{t} = \eta \boldsymbol{m}_{t-1} + (1-\eta) \boldsymbol{g}_{t}$ by applying momentum $\mu$ to the anti-gradient (the update vector is initialized according to $\boldsymbol{m}_0 = \{0\}^n$) and update the problem variable vector as follows:
 \begin{equation}
	\boldsymbol{c}_{t+1} = \boldsymbol{c}_t + \boldsymbol{m}_{t}\odot \mathbb{I}[|\boldsymbol{c}_t+\boldsymbol{m}_{t}|\leq 1]	
	\label{eq:sim}
\end{equation}
where $\mathbb{I}$ denotes the indicator function and  $\odot$ element-wise product. In other words, an update is applied to each element of $\boldsymbol{c}_t$ only if it does not cause it to to exceed the boundary of $[-1,1]$.  

These  iterations are repeated  $N$ times.  Subsequently, the solution of the original discrete problem (\ref{eq:qubo}) is calculated as its elementwise \emph{sign}. SimCIM is reminiscent to the Hopfield-Tank simulated annealer \cite{HopfieldTank}, but differs from it in the shape of the activation function.

The hyperparameters $\mu, \eta, \sigma$ are scalar values and relatively easy to tune. In contrast, $p_t$ is a discretized function of time, which poses a challenge to common hyperparameter optimization techniques due to the large dimensionality. 

\paragraph{Eigenvalue decomposition}

Since the matrix $J$ is real and symmetric, we can construct an eigenvalue decomposition $J = Q\Lambda Q^T,$ where $Q$ is an orthogonal matrix with the eigenvectors of matrix $J$ as its columns, and $\Lambda$ is a diagonal matrix with the eigenvalues of $J$ as diagonal elements $\Lambda_{ii}.$ 

With some simplifications $(\eta=\sigma=0, \ \boldsymbol{c}_{t} \ll \boldsymbol{1})$ the dynamics (\ref{eq:sim}) of the system can be described by the equation
$\boldsymbol{c}_{t+1} = \boldsymbol{c}_{t} + \mu (J \boldsymbol{c}_t - p_t \boldsymbol{c}_t).$
By performing eigenvalue decomposition and the change of variable $\boldsymbol{e} = Q^T\boldsymbol{c}$, the update equation simplifies to
$e_{t+1,i} = e_{t,i} + \mu (\Lambda_{ii} e_{t,i} - p_t e_{t,i})$ --- i.e.~the update is applied to individual elements of the vector $\boldsymbol{e}$.
Thus, when $p_t$ is greater than the highest eigenvalue of $J,$ both $\boldsymbol{e}$ and $\boldsymbol{c} = Q\boldsymbol{e}$ exponentially decay. Also, setting $p_t$ lower than all $\Lambda_{ii}$ will lead to exponential growth of all amplitudes $e_i$, and subsequent poor conversion of the iterations. Using these observations, we reparameterize the regularization function by introducing a normalized regularization function $\bar{p}_t$, which, as a rule, is restricted to the interval $[0,1]$:

\begin{equation}
p_t = \bar{p}_t  \left(\max_i \Lambda_{ii} - \min_i \Lambda_{ii}\right) + \min_i \Lambda_{ii}.
\end{equation} 

\paragraph{Choosing a learning rate}
 
To select the learning rate $\mu$ for each problem instance, we use an automatic procedure similar to the learning rate range test proposed in \cite{lrtest}. One optimization cycle of SimCIM (\ref{eq:sim}) with momentum $\eta=0$ is performed with an exponentially decaying $\mu$, starting from some high value. The learning rate is chosen at an iteration where the $l_1$ norm of gradient $\|\boldsymbol{g}_{t}\|_1$ starts to converge.  

\subsection{Reinforcement learning}

In reinforcement learning (RL) setting, an agent at each step $t$ interacts with some environment $\mathcal{E}$ by observing its state $s_t,$ performing an action $a_t$ sampled from its policy $\pi(a),$ and obtains a reward $r_t(s_t, a_t, s_{t+1}).$ One interaction session, called an episode, usually lasts until the agent reaches a terminal state or until the limit on the number of steps $T$ is reached. The goal of the agent is to maximize the expected sum of discounted rewards during the episode: $\mathbb{E_{\tau(\pi)}} \sum_{t=0}^T \gamma^{t} r_t(s_t, a_t, s_{t+1}),$ where $\tau(\pi)$ is a trajectory generated by the agent in the environment, and $\gamma \in (0,1]$ is the discount factor.

In actor-critic learning, an agent consists of two components. The actor, using observations from the environment, predicts the agent's policy $\pi(a).$ The actor's parameters are updated in the direction of improvement, which is estimated using sampled trajectories.  The critic predicts the value of each observation, which is then used to reduce the variance of actor's gradient. In the case of deep reinforcement learning, both actor and critic are usually  implemented as deep neural networks.

\section{Our approach}

We use a neural network based RL agent to control the scaled regularization function $\bar{p}_t.$ Every $m$ iterations of SimCIM, the agent observes the state of the optimization process and modifies $\bar{p}_t.$ Observations, actions and rewards from each SimCIM optimization cycle constitute one agent rollout: in an episode of $N$ SimCIM steps the agent performs $N/m$ actions.   

\paragraph{Actions}

The agent has a discrete action space: it is allowed to increase or decrease the regularization function, which is initialized at $\bar{p}_0=1.0$, by one of the values $\{ \pm p_\Delta, 0 \}.$ In addition to that, $\bar{p}_t$ is decreased by $\frac{m}{N}$ at each agent step to ensure that a random agent yields a decreasing regularization function. Between the agent steps, $\bar{p}_t$ is interpolated linearly; $\bar{p}_t$ is also clipped to the interval $[0,1.05]$ to limit the exploration area.

\paragraph{Observations}
The agent observes the current state of optimization variables in the eigenbasis of the problem matrix $J$, i.e. it is supplied with the set of amplitudes $\boldsymbol{e}_{t,i}$ (listed in the order of decreasing corresponding eigenvalues $\Lambda_{ii}$), as well as the elapsed time $t/N$ and the regularization function $\bar{p}_{t-1}$ from the previous step. The benefit of using  $\boldsymbol{e}_t$, rather than the actual amplitudes $\boldsymbol{c}_t$, as the state component, is that the former have a natural ordering according to the corresponding eigenvalues of $J$, while the components of $\boldsymbol{c}_t$ can be arbitrary permuted along with the rows and columns of $J$. This representation of the state therefore facilitates the transferability  of the agent across problems. 

To provide the agent with the information about the current problem instance for the purpose of transfer learning, we calculate problem features as $\boldsymbol{\phi}_j = \frac{1}{n}\sum_{i=1}^n |Q_{ij}|$. This means that $\boldsymbol{\phi}_j$ are scaled $l_1$ norms of the problem matrix eigenvectors. These features are ``static" observations that are fixed during the entire episode. Features $\boldsymbol{\phi}_j$ are provided to the agent at each step as a part of the observation.

\paragraph{Rewards}

In the case of combinatorial optimization, we are interested in finding solutions with the best quality (e.g. cut value) for each instance, while the path in which it has been reached is less important. Also, solutions with slightly different cut values may correspond to completely different bit configurations $\boldsymbol{x}$. Thus the current cut value or its difference between steps is not the best choice for the reward.

To address this issue, the Ranked Reward (R2) method was proposed in \cite{r2}. In R2, the environment maintains a list of discovered cut values $C_j$ for the last $P$ episodes (a ``leaderboard"), the $q$-th percentile $C^q$ is calculated over this list, and the new solution with the cut value $C$ is rewarded \emph{at the last step only} according to the rule
\begin{equation}
	\label{r2}
	r_{R2} = 
	\begin{cases}
	+1, & C > C^q \\
	-1, & C < C^q \\
	\pm 1 \ \  \text{randomly}, & C=C^q
	\end{cases},
\end{equation}
where $q$ and $P$ are hyperparameters.
This kind of reward ensures that the agent constantly improves its performance in search of better solutions. In the language of of self-play \cite{alphazero}, the agent is rewarded for beating most of its last results in a single-player game (being at the top of the leaderboard) and punished otherwise.

We propose a modification of this method that we dub Rescaled Ranked Rewards (R3) to account for imbalanced reward distribution:

\begin{equation}
	\label{r3}
	r_{R3} = 
	\begin{cases}
	+\frac{q}{100}, & C > C^q \\
	-(1-\frac{q}{100}), & C < C^q \\
	\bar{r}, & C=C^q
	\end{cases},
\end{equation}

where $\bar{r}$ is calculated in such a way that the average reward over the last $P$ episodes is equal to zero. This modification ensures that negative and positive rewards are balanced. It also ensures that solutions with $C > C^q$ are clearly distinguishable from  those with $C=C^q$, and hence discourages the agent from getting stuck in a local optimum.

\paragraph{Transfer learning}
\label{transfer}

The approach we propose requires training the agent for each problem instance separately, however it is possible to accelerate this process significantly by pre-training the agent on randomly generated problem instances.

We pre-train the agent on random adjacency matrices from the Erdős--Rényi distribution \cite{erdos} with a fixed connection probability of $0.06$. We select this value so that the pre-training distribution is close to that for the target set of problems. However, we observe that transfer works reliably for matrices with different structure, too. 

At each step of the pre-training process, the environment samples a new matrix $J,$ and the agent uses it to generate a batch of episodes and perform a gradient update. This is repeated a fixed number of times. Note that this procedure does not require any costly data labeling or using previously known solutions.

Once the training is complete, the agent is fine-funed in application to the specific problem of interest. This fine-tuning is performed in a similar manner: at each step the agent generates a batch of episodes using the matrix $J$ of the problem and performs a gradient update.

\paragraph{Implementation details}

The agent is implemented as two separate fully-connected networks (actor and critic) with two hidden layers of size 256 and $\tanh$ activation functions. These two networks take environment observation as input and produce policy and value function respectively.

The static features of the problem matrix $\boldsymbol{\phi}_j$ are not included in the network inputs; instead, they are used to calculate a set of parameters to perform feature-wise linear modulation (FiLM) \cite{film} of the last hidden layer in the actor network. The FiLM module is a linear layer that predicts a set of weights and biases that are used to scale and shift the activations of the actor's hidden layer element-wise.  

We train the agent using the PPO \cite{ppo} algorithm with $4$ epochs. The discount factor $\gamma$ is equal to $1.0$. SimCIM performs $N = 1000$ iterations per episode, and the agent acts every $m=10$ iterations, corresponding to $100$ steps per episode. The SimCIM algorithm allows efficient parallel implementation on a GPU, so we train the agent in batches of size $256$ (both for pre-training and fine-tuning). We use $q=0.99$ to calculate rewards in R2 and R3 methods; the leaderboard size $P$ is equal to 5 batch sizes for fine-tuning and one batch size for pre-training (since each problem instance is used to generate only one batch of episodes).  
The pre-training is performed for 30000 problem instances.

The SimCIM hyperparameters are chosen as follows. The momentum is set to $\eta=0.9$ and noise level to $\sigma=0.03.$ The learning rate $\mu$ is tuned automatically for each problem instance, including the random instances used for pre-training. The regularization function increment $p_\Delta$ is equal to $0.04.$

\section{Related Work}

Aside from classic heuristic methods for combinatorial optimization that can be found in industrial-scale packages like Gurobi~\cite{gurobi} and CPLEX~\cite{cplex}, many RL-based algorithms are emerging. Early works \cite{pointer, device} use RL to train recurrent neural networks with attention mechanisms to construct the solution iteratively. 
In later papers \cite{gn1, gn2, gn3, gn4, gn5, gn6} different kinds of graph neural networks are used in conjunction with RL to solve combinatorial problems on graphs by iteratively flipping bit values. 

In \cite{r2}, a permutation-invariant network was used as a reinforcement learning agent to solve the bin packing problem. This work introduced Ranked Reward to automatically control the learning curriculum of the agent.

Combining RL with heuristics was explored in \cite{combo}: one agent was used to select a subset of problem components, and another selected an heuristic algorithm to process them. 

In \cite{qaoa_rl}, a reinforcement learning agent was used to tune the parameters of a simulated quantum approximate optimization algorithm (QAOA) \cite{qaoa} to solve the Max-Cut problem and showed strong advantage over black-box parameter optimization methods on graphs with up to 22 nodes. QAOA was designed with near-term noisy quantum hardware in mind, however, at the current state of technology, the problem size is limited both in hardware and simulation.   

To the best of our knowledge, combining quantum-inspired algorithms with RL for combinatorial optimization in the context of practically significant problem sizes was not explored before.

\section{Experiments}

To evaluate our method, we use problem instances from Gset \cite{gset}, which is a set of graphs (represented by adjacency matrices $J$) that is commonly used to benchmark Max-Cut solvers. Gset contains problems of practically significant sizes, from hundreds to thousands of variables from several different distributions. 

We concentrate on graphs G1--10. Of these, G1--G5 appear to belong to the Erdős--Rényi \cite{erdos} model with the connection probability approximately equal to 0.06, while G6--G10 are weighted graphs with the same adjacency structure, but with approximately half of the edges having weights equal to $-1$. All of these graphs have 800 nodes.

For all our experiments, we use a single machine with a GeForce RTX 2060 GPU.

\subsection{Performance}

The agent, pre-trained and fine-tuned as described in Section~\ref{transfer}, is used to generate a batch of solutions, for which we calculate the maximum and median cut value. We also report the fraction of solved instances: the problem is considered solved if the maximum cut over the batch is equal to the best known value reported in \cite{bls}.

The results are presented in Table \ref{tab:res}. The obtained maximum and median are normalized by this best known value; the normalized values are further averaged over instances G1--G10 and over three random seeds for each instance (for each random seed we pre-train a new agent). Note that problem instances G6--G10 belong to a distribution never seen by the agent during the pre-training.

We compare our method to two baseline approaches to tuning the regularization function of SimCIM. In the first approach (labelled ``Linear"), the scaled regularization function $\bar{p}_t$ is decaying linearly from 1 to 0 during the $N$ SimCIM iterations; in our reinforcement learning setting, this is equivalent to the agent that always chooses zero increment as the action. In the second approach (labelled ``Manual"), which has been used in the original SimCIM paper \cite{sim}, the regularization function is a parameterized hyperbolic tangent function:
\begin{equation}
    p_t = J_m O (\tanh (S(t/N-0.5)) + D),
    \label{eq:tanh}
\end{equation}
where $J_m = \max_i \sum_j |J_{ij}|; \ $ $t/N$ is a normalized iteration number and $O, S, D$ are the scale and shift parameters. 
These parameters are tuned manually for all instances G1--G10 at once. If manually tuned in this fashion, SimCIM   solves 8 of G1--G10 instances, however the result is stochastic and the probability of solving each instance is different \cite{sim}. We evaluate the baselines by sampling 30 batches of solutions (batch size 256) for each instance and averaging the statistics (maximum, median, fraction of solved) over all batches of all instances.

We also compare our approach to a well-known evolutionary algorithm CMA-ES \cite{cmaes} (population size 10). We parameterize the regularization function for iteration $t$ according to Eq.~(\ref{eq:tanh}), and CMA-ES is used to tune $D \in [-3,3]$ and $O, S \in [0.01, 10]$ (exponential scale) for at most 500 SimCIM evaluations in batches of size 256 each. We maximize $\mathcal{C}_{max} + q_{max},$ where $\mathcal{C}_{max}$ is the maximum cut over the batch, and $q_{max}$ is the fraction of values in the batch equal to $\mathcal{C}_{max}.$ Since all elements of $J$ are integer, so is the cut value, while $0 < q_{max} \leq 1$. As a result, this objective orders batches first by the maximum cut value, and then by the probability to obtain it. After the optimization is finished, the best parameters are selected, and a new batch of solutions is sampled with these parameters. We report results from the batches obtained in this manner, averaged over three random seeds and over all instances.

\begin{table*}[t]
\caption{Performance on Gset: maximum and median normalized cut values are averaged over the instances (G1--G10); Agent-$K$ denotes an agent fine-tuned for $K$ episodes, Agent-0 is not fine-tuned. Standard deviation over three random seeds is reported in brackets for each value. }
\label{tab:res}
\vskip 0.1in
\begin{center}
\begin{scriptsize}
\begin{sc}
\begin{tabular}{l|lll|llll}
\toprule
{} &  Linear &  Manual &  CMA-ES & Agent-0 & Agent-100 & Agent-200 & Agent-500 \\
\midrule
Maximum &  0.9993 (2e-05) &  0.9997 (2e-05) &  0.9995 (8e-05) &  0.9990 (2e-04) &  0.9996 (8e-05) &  0.9997 (1e-05) &  \textbf{0.9998} (0e+00) \\
Median  &  0.9942 (5e-05) &  0.9946 (3e-04) &  0.9933 (4e-04) &  0.9901 (2e-03) &  0.9901 (1e-04) &  0.9925 (2e-03) &  \textbf{0.9979} (4e-04) \\
Solved  &  0.2000 (0e+00) &  0.6667 (5e-02) &  0.6000 (0e+00) &  0.1333 (5e-02) &  0.6000 (8e-02) &  0.7333 (5e-02) &  \textbf{0.8000} (0e+00) \\
\bottomrule
\end{tabular}
\end{sc}
\end{scriptsize}
\end{center}
\vskip -0.1in
\end{table*}

Though the pre-trained agent without fine-tuning (Agent-0) is even worse than the baselines, fine-tuning rapidly improves the performance of the agent.
The fine-tuned agent does not solve all instances in G1--G10, however it discovers high-quality solutions more reliably than the benchmarks.

CMA-ES is capable of solving each of G1--G10 instances: we observed that the best known value appeared at least once for each instance during several trials with different seeds. However, for some instances this result is not reproducible due to the stochastic nature of SimCIM: a new batch of solutions generated with the best parameters found by CMA-ES may yield a lower maximum cut. In this sense, the results for CMA-ES are worse than for the manually tuned baseline.

\begin{figure}[ht]
\vskip 0.2in
\begin{center}
  \centerline{\includegraphics[width=\columnwidth]{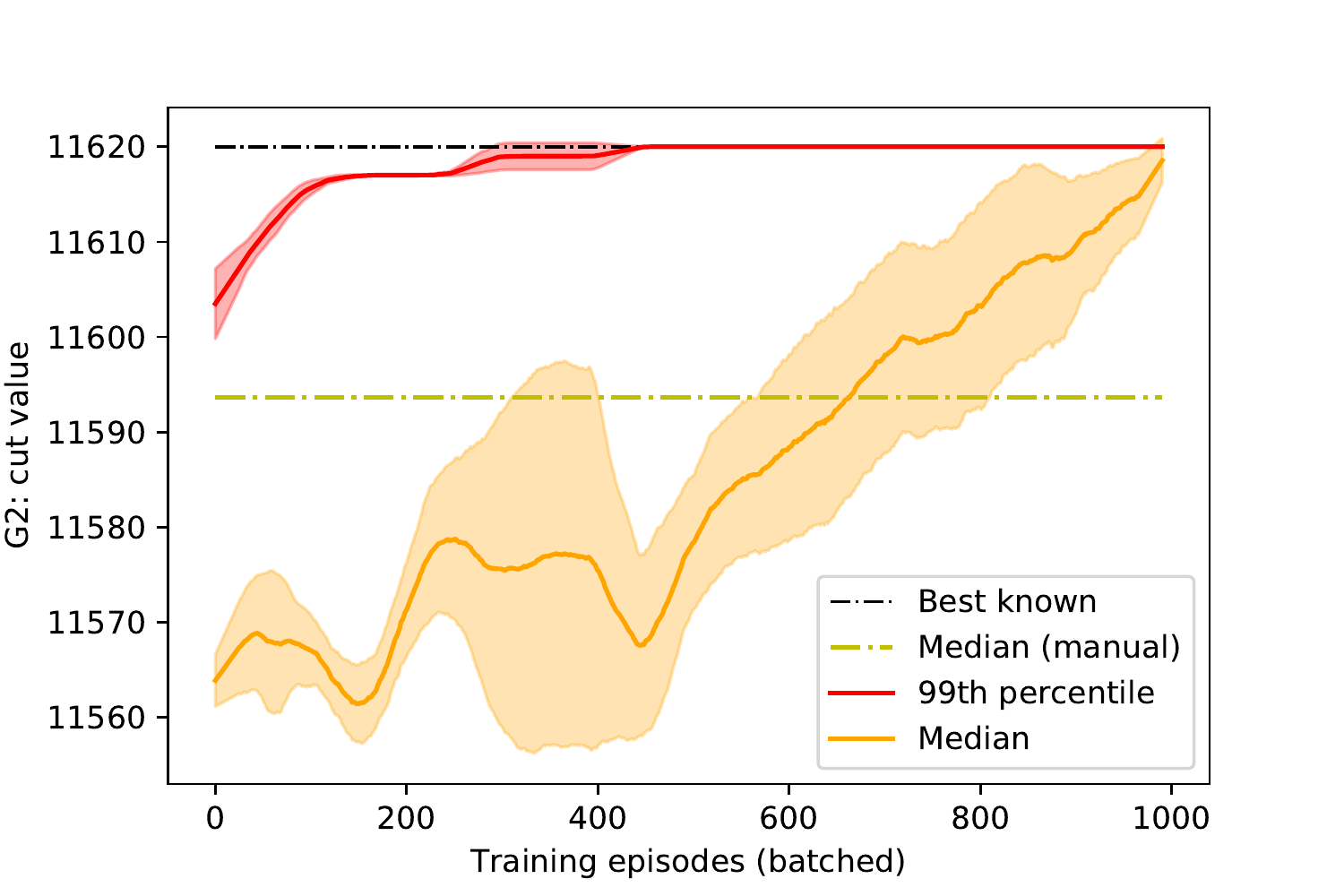}}
  \caption{Example: dynamics of the cut value obtained on G2 during fine-tuning, standard deviation is calculated over three random seeds (smoothed with Savitzky--Golay filter).}
  \label{fig:median}
 \end{center}
\vskip -0.2in
\end{figure}

Figure \ref{fig:median} demonstrates the dynamics of the maximum and median cut values for the G2 instance during the process of fine-tuning. The median value continues to improve, even after the agent has found the best known value, and eventually surpasses the manually tuned baseline. This means that the agent still finds new ways to reach solutions with the best known cut. A further advantage of our agent is that it adaptively optimizes the regularization hyperparameter during the test run by taking the current trajectories $\boldsymbol{c}_t$ into account.

The exact maximum cut values after fine-tuning and best know solutions for specific instances G1--G10 are presented in Table \ref{tab:bls}. We see that the agent stably finds the best known solutions for G1--G8 and closely lying solutions for G9--G10. The reason it fails to solve G9 and G10 is that the policy found by the agent corresponds to a deep local optimum that the agent is unable to escape by gradient descent. In contrast, CMA-ES does not use gradient descent and is focused on exploratory search in a broad range of parameters, and hence is sometimes able to  solve these graphs. However, even with CMA-ES, the solution probability is vanishingly small: $1.3 \times 10^{-5}$ for G9 and $9.8 \times 10^{-5}$ for G10. 

We also note the difference in the numbers of samples used by the automatic methods --- our agent and CMA-ES --- as compared to the manual hyperparameter tuning and the linear variation of the hyperparameter. In the former case, the total number of samples consumed including both training (fine-tuning) and at test equalled $\sim 256\times 500=128000$. On the other hand, the manual tuning required much fewer samples (tens of thousands), while the linear setting did not involve any tuning at all. Hence it is fair to say that the linear and manual methods are much more sample-efficient.

\begin{table*}[t]
\caption{Results for specific Gset instances: best known cut value, best value obtained by the agent, their difference and the probability for the fully trained agent to find a solution corresponding to its best value.}
\label{tab:bls}
\vskip 0.1in
\begin{center}
\begin{small}
\begin{sc}
\begin{tabular}{l|rrrrrrrrrr}
\toprule
{} &     G1 &     G2 &     G3 &     G4 &     G5 &    G6 &    G7 &    G8 &    G9 &   G10 \\
\midrule
Best \cite{bls} &  11624 &  11620 &  11622 &  11646 &  11631 &  2178 &  2006 &  2005 &  2054 &  2000 \\
Agent          &  11624 &  11620 &  11622 &  11646 &  11631 &  2178 &  2006 &  2005 &  2050 &  1999 \\
Difference          &      0 &      0 &      0 &      0 &      0 &     0 &     0 &     0 &    -4 &    -1 \\
Probability & 0.87 & 0.49 & 0.81 & 0.93 & 0.34 & 0.53 & 0.82 & 0.92 & 0.61 & 0.46 \\
\bottomrule
\end{tabular}
\end{sc}
\end{small}
\end{center}
\vskip -0.1in
\end{table*}

\subsection{Ablation study}  

\begin{figure}[ht]
\vskip 0.2in
\begin{center}
  \centerline{\includegraphics[width=\columnwidth]{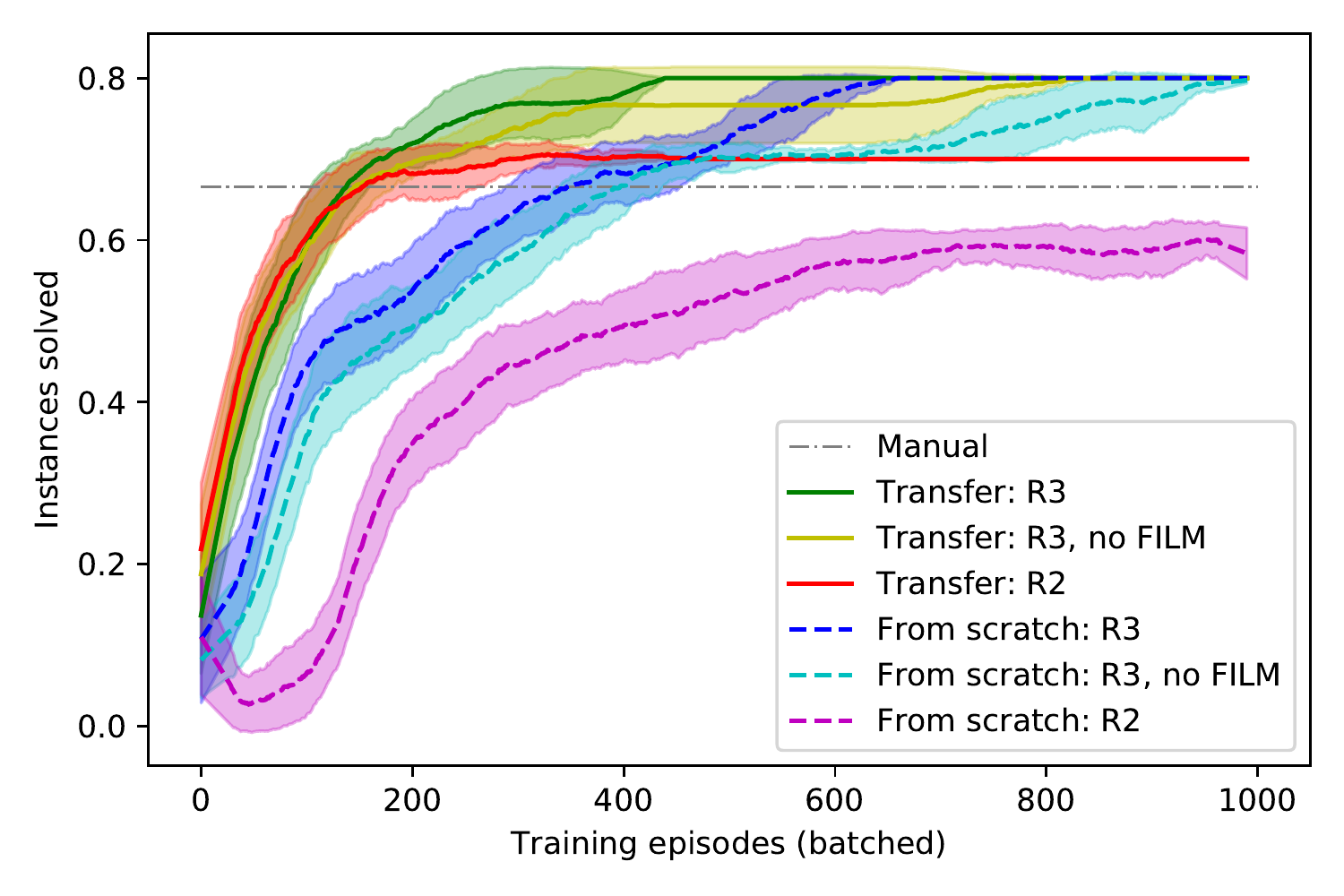}}
  \caption{Ablation study: averaged fraction of solved problem instances versus the number of episodes of fine-tuning for each instance (smoothed with Savitzky--Golay filter). Standard deviation is calculated over three random seeds. ``Transfer" and ``From scratch" are used to denote the agent with and without pre-training, respectively.}
  \label{fig:abl}
  \end{center}
\vskip -0.2in
\end{figure}

We study the effect of the three main components of our approach: transfer learning from random problems, Rescaled Ranked Rewards (R3) scheme, and feature-wise linear modulation (FiLM) of the actor network with the problem features.
\begin{itemize}
    \item To study the effect of the policy transfer, we train pairs of agents with the same hyperparameters, architecture and reward type, but with and without pre-training on randomly sampled problems. In the latter case, the parameters of the agent are initialized randomly.
    \item We compare our R3 method with the original R2 method both with and without pre-training. 
    \item We study the effect of FiLM  by removing the static observations extracted from the problem matrix $J$ from the observation and the FiLM layer from the agent.
\end{itemize}
We report the fraction of solved problems, averaged over instances G1--G10 and over three random seeds for each instance. The results are presented in Table \ref{tab:abl} and Fig.~\ref{fig:abl}.

\begin{table*}[t]
\caption{Ablation study, fraction of problems solved. Agent-$K$ denotes an agent fine-tuned for $K$ episodes. Standard deviation over three random seeds is reported in brackets for each value.}
\label{tab:abl}
\vskip 0.1in
\begin{center}
\begin{small}
\begin{sc}
\begin{tabular}{l|ccc|ccc}
\toprule
{} &  &      Pre-training       &       &  &     No pre-training        &       \\
Solved &       R3 & R3, no FiLM &    R2 &           R3 & R3, no FiLM &    R2 \\
\midrule
Solved (100 it.) &  0.60 (8e-02) &  \textbf{0.63} (5e-02) &  0.60 (8e-02) &  0.40 (0e+00) &  0.37 (5e-02) &  0.10 (8e-02) \\
Solved (200 it.) &  \textbf{0.73} (5e-02) &  0.70 (0e+00) &  0.67 (5e-02) &  0.47 (5e-02) &  0.53 (9e-02) &  0.33 (5e-02) \\
Solved (500 it.) &  \textbf{0.80} (0e+00) &  0.77 (5e-02) &  0.70 (0e+00) &  0.73 (5e-02) &  0.73 (5e-02) &  0.53 (5e-02) \\
\bottomrule
\end{tabular}
\end{sc}
\end{small}
\end{center}
\vskip -0.1in
\end{table*}

According to the results, all of the above listed features are essential for the agent's performance. We see, in particular, that the pre-trained agent with both FiLM and R3 rewards experiences a slightly slower start, but eventually finds better optima faster than ablated agents. 

\subsection{Rescaled ranked rewards}
\label{sec:an}

\begin{figure}[ht]
\vskip 0.2in
\begin{center}
  \centerline{\includegraphics[width=\columnwidth]{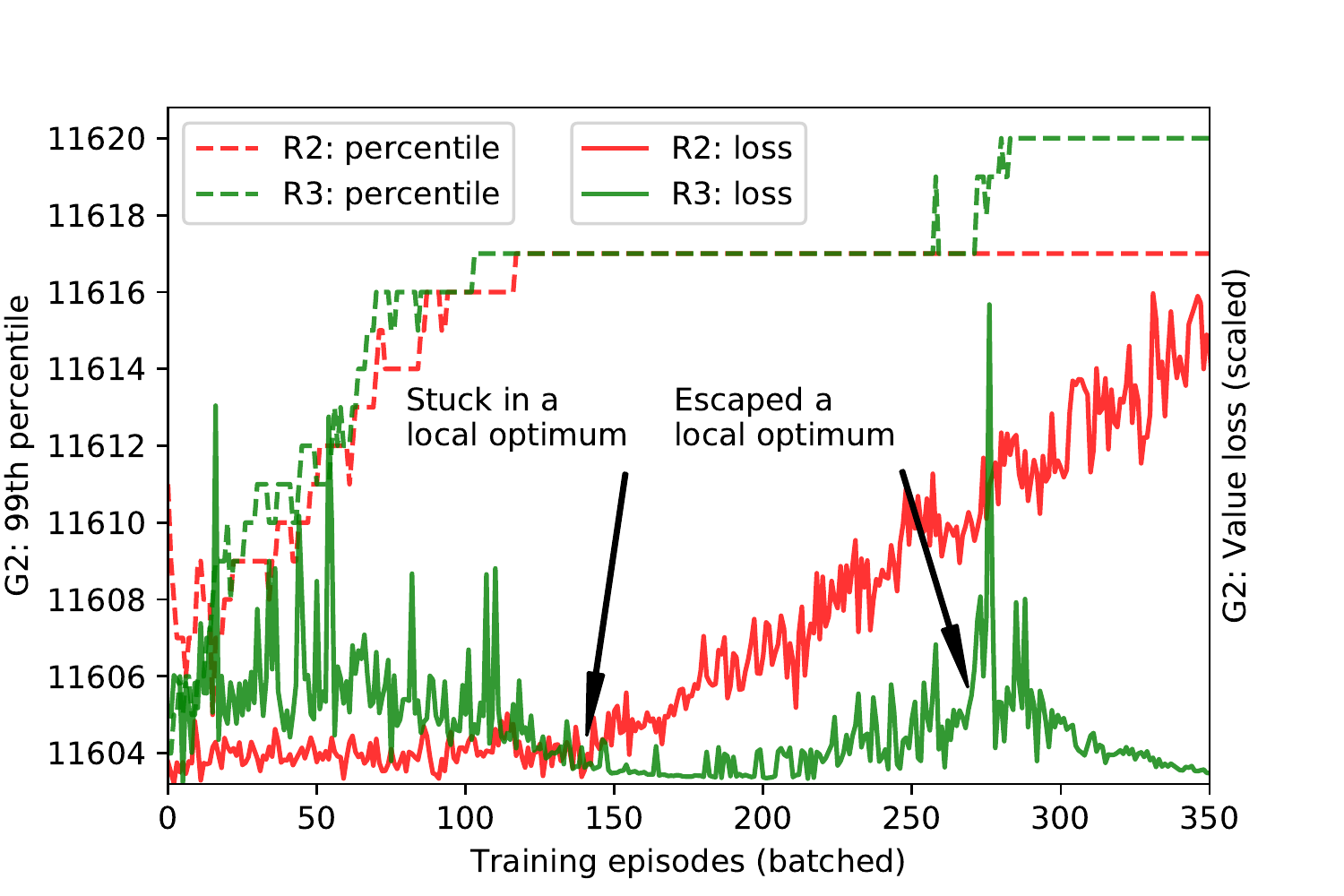}}
  \caption{Value loss and 99th percentile during fine-tuning on G2 for R2 and R3  when dealing with local optima.}
  \label{fig:loss}
\end{center}
\vskip -0.2in
\end{figure}

The analysis of specific problem instances helps to demonstrate the advantage of the R3 method. We analyze the behavior of the 99-th percentile of the solution cut values (the one used to distribute rewards in R2 and R3) on the G2 instance from Gset in Fig.~\ref{fig:loss}. 
G2 has several local optima with the same cut value 11617, which are relatively easy to reach. 
When the agent is stuck in a local optimum, many solutions generated by the agent are likely to have their cut values equal to the percentile, while solutions with higher cut values may appear infrequently. 

In the R2 scheme (\ref{r2}), the agent gets random $\pm 1$ rewards for local-optimum solutions and $+1$ for better ones. Thus infrequent solutions with higher cut values become almost indistinguishable from the local-optimum solutions. Furthermore, the fraction of episodes with local-optimum solutions increases, which results in a large fraction of random rewards, thereby preventing the efficient training of the critic network. This is evident from the monotonic growth of the value loss function in Fig.~\ref{fig:loss}.

In our R3 scheme (\ref{r3}), in contrast, the rewards for the local-optimum solutions are deterministic and dependent on the frequency of such solutions. The more often the agent reaches them, the lower the reward, while the reward for solutions with higher cut values is fixed. Eventually, better solutions outweigh sub-optimal ones, and the agent escapes the local optimum. This moment is indicated by a significant increase of the value loss: the agent starts exploring new, more promising states.

\section{Discussion and future work}
One of the benefits of our approach is the lightweight architecture of our agent, which allows efficient GPU implementation along with the SimCIM algorithm itself. This allows us to rapidly fine-tune the agent for each problem instance. 
However, the fully-connected architecture makes it harder to apply our pre-trained agent to problems of various sizes, since the size of the network input layer depends on the problem size.
Hence it would be interesting to explore using size-agnostic architectures for the agent, like graph neural networks.

Another future research direction is to train the agent to vary more SimCIM hyperparameters, such as the scaling of the adjacency matrix or the noise level. Additionally, it would be interesting to explore using meta-learning at the pre-training step to accelerate the fine-tuning process.

Lastly, with our approach, each novel instance requires a new   run of fine-tuning, leading to a large number of required samples compared with simple instance-agnostic heuristics. In order to make our approach viable from a practical point of view, we hope to address generalization across different, novel, problem instances more efficiently.

\section{Conclusion}

In this work we proposed an RL-based approach to tuning the regularization function of SimCIM, a quantum-inspired algorithm, to robustly solve the Ising problem. Our hybrid approach shows strong advantage over heuristics and a black-box approach, and allows us to sample high-quality solutions with high probability.

We proposed an improvement over the Ranked Reward (R2) scheme, called Rescaled Ranked Reward (R3), which allows the agent to constantly improve the current solution while avoiding local optima. We also demonstrated that our algorithm may be accelerated significantly by pre-training the agent on randomly generated problem instances, while being able to generalize to out-of-distribution problems.

Importantly, our approach is not limited to SimCIM or even the Ising problem, but can be readily generalised to any algorithm based on continuous relaxation of discrete optimisation.

\section*{Acknowledgements}

We would like to thank Egor Tiunov for providing the manual tuning data and William Clements and Vitaly Kurin for helpful discussions. This project has received funding from the Russian Science
Foundation (19-71-10092).

\bibliography{example_paper}

\begin{thebibliography}{36}
\providecommand{\natexlab}[1]{#1}
\providecommand{\url}[1]{\texttt{#1}}
\expandafter\ifx\csname urlstyle\endcsname\relax
  \providecommand{\doi}[1]{doi: #1}\else
  \providecommand{\doi}{doi: \begingroup \urlstyle{rm}\Url}\fi

\bibitem[cpl(2019)]{cplex}
Cplex optimizer.
\newblock \url{https://www.ibm.com/analytics/cplex-optimizer}, 2019.

\bibitem[gur(2019)]{gurobi}
Gurobi optimization.
\newblock \url{https://www.gurobi.com/}, 2019.

\bibitem[Abe et~al.(2019)Abe, Xu, Sato, and Sugiyama]{gn5}
Abe, K., Xu, Z., Sato, I., and Sugiyama, M.
\newblock Solving np-hard problems on graphs by reinforcement learning without
  domain knowledge.
\newblock \emph{arXiv preprint arXiv:1905.11623}, 2019.

\bibitem[Barahona(1982)]{nphard}
Barahona, F.
\newblock On the computational complexity of ising spin glass models.
\newblock \emph{Journal of Physics A: Mathematical and General}, 15\penalty0
  (10):\penalty0 3241, 1982.

\bibitem[Barrett et~al.(2019)Barrett, Clements, Foerster, and Lvovsky]{gn6}
Barrett, T.~D., Clements, W.~R., Foerster, J.~N., and Lvovsky, A.
\newblock Exploratory combinatorial optimization with reinforcement learning.
\newblock \emph{arXiv preprint arXiv:1909.04063}, 2019.

\bibitem[Benlic \& Hao(2013)Benlic and Hao]{bls}
Benlic, U. and Hao, J.-K.
\newblock Breakout local search for the max-cut problem.
\newblock \emph{Engineering Applications of Artificial Intelligence},
  26\penalty0 (3):\penalty0 1162--1173, 2013.

\bibitem[Dumoulin et~al.(2016)Dumoulin, Shlens, and Kudlur]{film}
Dumoulin, V., Shlens, J., and Kudlur, M.
\newblock A learned representation for artistic style.
\newblock \emph{arXiv preprint arXiv:1610.07629}, 2016.

\bibitem[Erd{\H{o}}s \& R{\'e}nyi(1960)Erd{\H{o}}s and R{\'e}nyi]{erdos}
Erd{\H{o}}s, P. and R{\'e}nyi, A.
\newblock On the evolution of random graphs.
\newblock \emph{Publ. Math. Inst. Hung. Acad. Sci}, 5\penalty0 (1):\penalty0
  17--60, 1960.

\bibitem[Farhi et~al.(2014)Farhi, Goldstone, and Gutmann]{qaoa}
Farhi, E., Goldstone, J., and Gutmann, S.
\newblock A quantum approximate optimization algorithm.
\newblock \emph{arXiv preprint arXiv:1411.4028}, 2014.

\bibitem[Feurer \& Hutter(2018)Feurer and Hutter]{hpo}
Feurer, M. and Hutter, F.
\newblock Hyperparameter optimization.
\newblock In Hutter, F., Kotthoff, L., and Vanschoren, J. (eds.),
  \emph{Automatic Machine Learning: Methods, Systems, Challenges}, pp.\  3--38.
  Springer, 2018.
\newblock In press, available at http://automl.org/book.

\bibitem[Hansen et~al.(2003)Hansen, M{\"u}ller, and Koumoutsakos]{cmaes}
Hansen, N., M{\"u}ller, S.~D., and Koumoutsakos, P.
\newblock Reducing the time complexity of the derandomized evolution strategy
  with covariance matrix adaptation (cma-es).
\newblock \emph{Evolutionary computation}, 11\penalty0 (1):\penalty0 1--18,
  2003.

\bibitem[Hopfield \& Tank(1986)Hopfield and Tank]{HopfieldTank}
Hopfield, J. and Tank, D.
\newblock Computing with neural circuits: a model.
\newblock \emph{Science}, 233\penalty0 (4764):\penalty0 625--633, 1986.
\newblock ISSN 0036-8075.
\newblock \doi{10.1126/science.3755256}.
\newblock URL \url{https://science.sciencemag.org/content/233/4764/625}.

\bibitem[Inagaki et~al.(2016)Inagaki, Haribara, Igarashi, Sonobe, Tamate,
  Honjo, Marandi, McMahon, Umeki, Enbutsu, et~al.]{cim1}
Inagaki, T., Haribara, Y., Igarashi, K., Sonobe, T., Tamate, S., Honjo, T.,
  Marandi, A., McMahon, P.~L., Umeki, T., Enbutsu, K., et~al.
\newblock A coherent ising machine for 2000-node optimization problems.
\newblock \emph{Science}, 354\penalty0 (6312):\penalty0 603--606, 2016.

\bibitem[Khairy et~al.(2019)Khairy, Shaydulin, Cincio, Alexeev, and
  Balaprakash]{qaoa_rl}
Khairy, S., Shaydulin, R., Cincio, L., Alexeev, Y., and Balaprakash, P.
\newblock Learning to optimize variational quantum circuits to solve
  combinatorial problems, 2019.

\bibitem[Khalil et~al.(2017)Khalil, Dai, Zhang, Dilkina, and Song]{gn1}
Khalil, E., Dai, H., Zhang, Y., Dilkina, B., and Song, L.
\newblock Learning combinatorial optimization algorithms over graphs.
\newblock In \emph{Advances in Neural Information Processing Systems}, pp.\
  6348--6358, 2017.

\bibitem[King et~al.(2018)King, Bernoudy, King, Berkley, and Lanting]{nmfa}
King, A.~D., Bernoudy, W., King, J., Berkley, A.~J., and Lanting, T.
\newblock Emulating the coherent ising machine with a mean-field algorithm.
\newblock \emph{arXiv preprint arXiv:1806.08422}, 2018.

\bibitem[Kirkpatrick et~al.(1983)Kirkpatrick, Gelatt, and Vecchi]{annealing}
Kirkpatrick, S., Gelatt, C.~D., and Vecchi, M.~P.
\newblock Optimization by simulated annealing.
\newblock \emph{Science}, 220\penalty0 (4598):\penalty0 671--680, 1983.

\bibitem[Kool et~al.(2018)Kool, van Hoof, and Welling]{gn3}
Kool, W., van Hoof, H., and Welling, M.
\newblock Attention, learn to solve routing problems!
\newblock \emph{arXiv preprint arXiv:1803.08475}, 2018.

\bibitem[Laterre et~al.(2018)Laterre, Fu, Jabri, Cohen, Kas, Hajjar, Dahl,
  Kerkeni, and Beguir]{r2}
Laterre, A., Fu, Y., Jabri, M.~K., Cohen, A.-S., Kas, D., Hajjar, K., Dahl,
  T.~S., Kerkeni, A., and Beguir, K.
\newblock Ranked reward: Enabling self-play reinforcement learning for
  combinatorial optimization.
\newblock \emph{arXiv preprint arXiv:1807.01672}, 2018.

\bibitem[Leleu et~al.(2019)Leleu, Yamamoto, McMahon, and Aihara]{destab}
Leleu, T., Yamamoto, Y., McMahon, P.~L., and Aihara, K.
\newblock Destabilization of local minima in analog spin systems by correction
  of amplitude heterogeneity.
\newblock \emph{Physical review letters}, 122\penalty0 (4):\penalty0 040607,
  2019.

\bibitem[Li et~al.(2018)Li, Chen, and Koltun]{gn2}
Li, Z., Chen, Q., and Koltun, V.
\newblock Combinatorial optimization with graph convolutional networks and
  guided tree search.
\newblock In \emph{Advances in Neural Information Processing Systems}, pp.\
  539--548, 2018.

\bibitem[Marzec(2016)]{portfolio2}
Marzec, M.
\newblock Portfolio optimization: applications in quantum computing.
\newblock \emph{Handbook of High-Frequency Trading and Modeling in Finance
  (John Wiley \& Sons, Inc., 2016) pp}, pp.\  73--106, 2016.

\bibitem[McGeoch et~al.(2019)McGeoch, Harris, Reinhardt, and Bunyk]{dwave}
McGeoch, C.~C., Harris, R., Reinhardt, S.~P., and Bunyk, P.~I.
\newblock Practical annealing-based quantum computing.
\newblock \emph{Computer}, 52\penalty0 (6):\penalty0 38--46, 2019.

\bibitem[McMahon et~al.(2016)McMahon, Marandi, Haribara, Hamerly, Langrock,
  Tamate, Inagaki, Takesue, Utsunomiya, Aihara, et~al.]{cim2}
McMahon, P.~L., Marandi, A., Haribara, Y., Hamerly, R., Langrock, C., Tamate,
  S., Inagaki, T., Takesue, H., Utsunomiya, S., Aihara, K., et~al.
\newblock A fully programmable 100-spin coherent ising machine with all-to-all
  connections.
\newblock \emph{Science}, 354\penalty0 (6312):\penalty0 614--617, 2016.

\bibitem[Mirhoseini et~al.(2017)Mirhoseini, Pham, Le, Steiner, Larsen, Zhou,
  Kumar, Norouzi, Bengio, and Dean]{device}
Mirhoseini, A., Pham, H., Le, Q.~V., Steiner, B., Larsen, R., Zhou, Y., Kumar,
  N., Norouzi, M., Bengio, S., and Dean, J.
\newblock Device placement optimization with reinforcement learning.
\newblock In \emph{Proceedings of the 34th International Conference on Machine
  Learning-Volume 70}, pp.\  2430--2439. JMLR. org, 2017.

\bibitem[Mittal et~al.(2019)Mittal, Dhawan, Medya, Ranu, and Singh]{gn4}
Mittal, A., Dhawan, A., Medya, S., Ranu, S., and Singh, A.
\newblock Learning heuristics over large graphs via deep reinforcement
  learning.
\newblock \emph{arXiv preprint arXiv:1903.03332}, 2019.

\bibitem[Perdomo-Ortiz et~al.(2012)Perdomo-Ortiz, Dickson, Drew-Brook, Rose,
  and Aspuru-Guzik]{protein}
Perdomo-Ortiz, A., Dickson, N., Drew-Brook, M., Rose, G., and Aspuru-Guzik, A.
\newblock Finding low-energy conformations of lattice protein models by quantum
  annealing.
\newblock \emph{Scientific reports}, 2:\penalty0 571, 2012.

\bibitem[Schulman et~al.(2017)Schulman, Wolski, Dhariwal, Radford, and
  Klimov]{ppo}
Schulman, J., Wolski, F., Dhariwal, P., Radford, A., and Klimov, O.
\newblock Proximal policy optimization algorithms.
\newblock \emph{arXiv preprint arXiv:1707.06347}, 2017.

\bibitem[Silver et~al.(2017)Silver, Hubert, Schrittwieser, Antonoglou, Lai,
  Guez, Lanctot, Sifre, Kumaran, Graepel, et~al.]{alphazero}
Silver, D., Hubert, T., Schrittwieser, J., Antonoglou, I., Lai, M., Guez, A.,
  Lanctot, M., Sifre, L., Kumaran, D., Graepel, T., et~al.
\newblock Mastering chess and shogi by self-play with a general reinforcement
  learning algorithm.
\newblock \emph{arXiv preprint arXiv:1712.01815}, 2017.

\bibitem[Smith(2017)]{lrtest}
Smith, L.~N.
\newblock Cyclical learning rates for training neural networks.
\newblock In \emph{2017 IEEE Winter Conference on Applications of Computer
  Vision (WACV)}, pp.\  464--472. IEEE, 2017.

\bibitem[Tiunov et~al.(2019)Tiunov, Ulanov, and Lvovsky]{sim}
Tiunov, E.~S., Ulanov, A.~E., and Lvovsky, A.
\newblock Annealing by simulating the coherent ising machine.
\newblock \emph{Optics express}, 27\penalty0 (7):\penalty0 10288--10295, 2019.

\bibitem[Ulanov et~al.(2019)Ulanov, Tiunov, and Lvovsky]{ulanov2019}
Ulanov, A.~E., Tiunov, E.~S., and Lvovsky, A.
\newblock Quantum-inspired annealers as boltzmann generators for machine
  learning and statistical physics.
\newblock \emph{arXiv preprint arXiv:1912.08480}, 2019.

\bibitem[Venturelli \& Kondratyev(2019)Venturelli and Kondratyev]{portfolio1}
Venturelli, D. and Kondratyev, A.
\newblock Reverse quantum annealing approach to portfolio optimization
  problems.
\newblock \emph{Quantum Machine Intelligence}, 1\penalty0 (1-2):\penalty0
  17--30, 2019.

\bibitem[Vinyals et~al.(2015)Vinyals, Fortunato, and Jaitly]{pointer}
Vinyals, O., Fortunato, M., and Jaitly, N.
\newblock Pointer networks.
\newblock In \emph{Advances in Neural Information Processing Systems}, pp.\
  2692--2700, 2015.

\bibitem[Xinyun \& Yuandong(2018)Xinyun and Yuandong]{combo}
Xinyun, C. and Yuandong, T.
\newblock Learning to perform local rewriting for combinatorial optimization.
\newblock \emph{arXiv preprint arXiv:1810.00337}, 2018.

\bibitem[Ye(2003)]{gset}
Ye, Y.
\newblock Gset max-cut problem set.
\newblock \url{https://web.stanford.edu/~yyye/yyye/Gset/}, 2003.

\end{thebibliography}
\bibliographystyle{icml2020}

\end{document}